\documentclass[letterpaper]{article}

\usepackage{natbib,alifeconf}  
\usepackage{algorithm2e}
\usepackage{xcolor}

\newcommand{\eg}{{\em e.g.}}
\newcommand{\ie}{{\em i.e.}}

\title{The Environmental Discontinuity Hypothesis for Down-Sampled Lexicase Selection}

\author{Ryan Boldi$^{1}$, Thomas Helmuth$^{2}$ \and Lee Spector$^{3, 1}$ \\
\mbox{}\\
$^1$University of Massachusetts, Amherst, MA 01003\\
$^2$Hamilton College, Clinton, NY 13323 \\
$^3$Amherst College, Amherst, MA 01002 \\
rbahlousbold@umass.edu}

\begin{document}
\maketitle

\begin{abstract}
  Down-sampling training data has long been shown to improve the generalization performance of a wide range of machine learning systems. Recently, down-sampling has proved effective in genetic programming (GP) runs that utilize the lexicase parent selection technique. Although this down-sampling procedure has been shown to significantly improve performance across a variety of problems, it does not seem to do so due to encouraging adaptability through environmental change. We hypothesize that the random sampling that is performed every generation causes discontinuities that result in the population being unable to adapt to the shifting environment. We investigate modifications to down-sampled lexicase selection in hopes of promoting incremental environmental change to scaffold evolution by reducing the amount of jarring discontinuities between the environments of successive generations. In our empirical studies, we find that forcing incremental environmental change is not significantly better for evolving solutions to program synthesis problems than simple random down-sampling. In response to this, we attempt to exacerbate the hypothesized prevalence of discontinuities by using only disjoint down-samples to see if it hinders performance. We find that this also does not significantly differ from the performance of regular random down-sampling. These negative results raise new questions about the ways in which the composition of sub-samples, which may include synonymous cases, may be expected to influence the performance of machine learning systems that use down-sampling.
\end{abstract}

\section{Introduction}


Genetic Programming (GP) is a supervised learning technique that takes inspiration from evolution to create computer programs that solve a variety of problems. In GP, the specifications for a program are defined in terms of input-output cases. Mimicking an evolutionary process, a set of initial programs are created at random. Then, the programs are evaluated on the input cases and compared to the ground truth output value to generate error values for each individual. A set of these programs that GP deems appropriate are selected, mutated, and sent to the next generation. This is repeated until a solution is found that passes all the training cases, or a limit is reached. There is a very important interaction at play during this process over the course of evolution. Namely, the interaction between the individuals in the population and the training cases that they might eventually be evaluated on. This interaction could roughly be thought of as the interaction between a creature and its environment during evolution. 

The selection step, one of the most important aspects of this evolutionary system, has been the subject of extensive study in GP and beyond. This is the step where individuals are picked from the current population to be parents for the offspring in the next generation. The selection step also dictates how many children each parent gets to have. In the past, methods performing this selection rely on aggregated fitness metrics \ie{ }they evaluate every individual on every test case, and sum the fitness across these test cases to generate a single fitness value. The individuals are then selected based on this single value using a selection method like fitness proportionate selection or tournament selection. Lexicase selection~\citep{helmuth2014solving} is a parent selection method that does not rely on aggregate fitness or performance metrics. Instead, individuals that are elite on a randomly shuffled ordering of the training cases are selected. This tends to result in the selection of specialist individuals that trade mediocre performance across all test cases for elite performance on a few \citep{helmuth2020importance}. It has empirically been shown to improve the performance of evolution across a wide range of problem domains. 

Using the analogy of a creature and its environment, lexicase selection can be thought of as using the random series of challenges that \emph{happen} to occur during a creature's life to dictate whether its genes survive to the next generation or not. Note that this seems very similar to what truly happens in evolution in nature: creatures must be able to survive the subset of life-threatening situations that \emph{happen} to occur during their lifetime. If a deadly situation that they are poor at avoiding never occurs, this negative performance never affects their selectability. Although lexicase selection seems to build a good analogy with biological evolution, it seems to be missing one key element: environmental change. This is because although each selected individual was ultimately elite on a possibly different set of cases, every single training case could be used every single generation. This is because of lexicase selection's random shuffling of the \emph{entire} training set. Every selection event (there is one selection event for every individual in the next population), a new random case ordering is used. Since the cases that come towards the beginning of a shuffle are generally more likely to be used in selection (as earlier cases are responsible for filtering out the vast majority of individuals in the selection pool), most (if not all) cases end up being used at least once every generation. This means that the environment for the entire population is generally constant from one generation to the next (although the individuals are exposed to a different set of cases for every selection).

\citet{Hernandez:2019:subsampling} recently proposed down-sampled lexicase selection as a method to reduce the number of individual evaluations performed by using a subset of the entire training set every generation. Beyond simply reducing computational effort, this method also seems to cause environmental change between generations as the set of cases that the individuals are exposed to changes over time. This results in cases that are not currently in the down-sample exerting no selection pressure on the population. In terms of the environment analogy, this can be thought of as the set of all possible challenges that an individual can face being constrained by where it is located. It is unlikely that the same creature would need to solve the problems that occur when climbing a tree \emph{and} swimming in the deep ocean. When using full lexicase selection, however, individuals could (and usually do) get tested on their ability to perform many combinations of tasks that are not constrained in any way over evolutionary time. Empirically, runs using down-sampled lexicase selection have been shown to significantly improve the performance of GP when compared to standard lexicase selection. However, \citet{Helmuth2021benefits} found evidence for the benefit of down-sampled lexicase selection \emph{not} arising from the environmental change it causes. Instead, the benefit might be due to an increased number of individuals being evaluated across evolution.

Although not found to be the driver behind down-sampled lexicase selection's success, evolutionary change has been found to be an important contributer to the evolvability of an evolutionary system \citep{levins_1968}. \citet{kastan2007varyingEnv} found that a varying environment facilitates faster evolutionary adaptation in biological simulations. This was later explored in using dynamic environments to improve the speed when performing Grammatical Evolution, a variant of GP, runs \citep{oneill2011dynamicEnv}. Changing environments has also been found to affect the speed and effecitveness of evolution in populations of \emph{Saccharomyces cerevisiae} yeast as well as digital organisms \citep{koning2019fluctuation}. However, the authors are not aware of any work attempting to cause gradual environmental change when performing GP runs with lexicase selection.

In this paper, we attempt to promote gradual environmental change in hopes of improving the performance of evolutionary runs. Promoting incremental change might have the effect of reducing the jarring discontinuities between the set of cases in down-samples for generations in close succession. This would make it easier for the population to adapt to the changing environment and ultimately result in more evolutionary success. First, we propose a variant of down-sampled lexicase selection, \emph{rolling down-sampled lexicase selection} that iteratively removes and adds cases to the sample over evolutionary time as an analogy to incremental environmental change in nature (\eg a river slowly cutting through a forest over many generations). We find that rolling with a step size of 1, or removing and adding one case at a time is not significantly better or worse than down-sampled lexicase selection at the 0.05 and 0.1 down-sample rate (5\% and 10\% of training cases in down-sample, respectively). 
Upon performing an experiment where we change the step size, we find more evidence that suggests that rolling is not better than simple random down-sampling. These results hint that either the random down-samples are indeed not jarringly different from each other when considering the behaviors they are selecting for, or that randomly rolling is not the way to promote environmental change.

In order to determine the extent to which the case differences between our down-samples are indeed jarring when randomly down-sampling, we attempt to move in the opposite direction of rolling. By doing this, we will be able determine whether large differences between the cases in successive down-samples affects the efficacy of our evolutionary runs. To do so, we use disjoint samples that would result in there being no identical cases in generations that come in close succession. With these disjoint samples, we aim to exacerbate the hypothesised jarring-ness that exists when using purely random samples to determine whether it is destructive to performance. GP runs using disjoint samples are found to not have a significant difference in performance when compared to simple random down-sampling.

Our experiments highlight the fact that the jarring differences between samples when randomly down-sampling do not have a significant detrimental effect on the success rate of GP runs. This is probably due to the large presence of synonymous cases in the training data. These synonymous cases mean that random down-samples are not actually jarringly different from each other in practice as although the case labels are entirely different, these cases can measure the same behavior and result in the same individual being selected. Although the alterations to down-sampled lexicase selection proposed in this paper do not improve on the solution rates of runs using this selection method, we find that the lack of a detrimental effect due to jarring discontinuities due to synonymous cases between the down-samples in practice to be a meaningful discovery that opens a promising new research direction: how does one leverage (or avoid) synonymous cases to maximise the success rate of GP runs using lexicase selection?



\section{Background and Related Work}
\subsection{Lexicase Selection}
Lexicase selection \citep{spector2012assessment, helmuth2014solving} is a parent selection technique that does not consider aggregate performance metrics to select parents for the next generation. Instead, lexicase selection selects individuals based on their performance on a random ordering of the training cases. First proposed to solve modal problems in GP, Lexicase selection has been applied in a vast range of domains, including rule based learning systems \citep{aenugu2019lexicase}, symbolic regression \citep{la2016epsilon}, machine learning \citep{la2020genetic, la2020learning, ding2022optimizing} and evolutionary robotics \citep{moore2017lex, moore2021objective, huizinga2018evolving}. Lexicase selection has been demonstrated to improve solution and generalization rates when compared to tournament selection and implicit fitness sharing \citep{helmuth2014solving}.

The lexicase selection algorithm, adapted from \citet{helmuth2020importance}, is outlined below:
\begin{enumerate}
    \item \textbf{candidates} is set to initially contain the entire population.
    \item \textbf{cases} is set to initially contain the entire training set shuffled in a random order.
    \item Collect individuals that have identical error vectors, and maintain only  one from each of these identical groups (for performance reasons).
    \item Until a parent is selected:
    \begin{enumerate}
        \item Remove all individuals from \textbf{candidates} that are not \emph{exactly} the best on the first case in \textbf{cases}.
        \item If only one individual remains in \textbf{candidates}, this becomes the selected parent.
        \item If there is only one case left in \textbf{cases}, pick an individual from \textbf{candidates} at random to become the parent.
        \item Else, remove the first case from \textbf{cases}.
    \end{enumerate}
\end{enumerate}

\subsection{Down-sampling Training Data}
Down-sampling has often been presented in the GP literature and beyond as a method to reduce computation and improve the generalization of evolutionary runs. \citet{ga94aGathercole} propose \emph{Dynamic Subset Selection}, where subsets of training data are picked to reduce the overall computational effort. \citet{Schmidt2005coevolving} co-evolve test cases and individuals, where the set of test cases every generation is smaller than the entire training set. Other methods of down-sampling are also prevalent in the field \citep{ Hmida:2019:ICDS, giacobini:ppsn2002:pp371, Martinez2014Subsampling}. Down-sampling is often used in Machine Learning in general to help improve generalization rates and reduce computational overhead of using large-scale datasets \citep{fatjon2021downsampling, Katharopoulos2018NotAS}.

\subsection{Down-sampled Lexicase Selection}
Down-sampled lexicase Selection, first proposed for expensive evolutionary robotics runs \citep{moore2017lex} and later formalized for GP \citep{Hernandez:2019:subsampling, Ofria:2019:GPTP}, is a method of decreasing the number of training cases that need to be evaluated every generation for lexicase selection. This results in parent selection requiring fewer total evaluations per generation. Instead of all individuals in the population being evaluated on all training cases to select parents, they are instead only evaluated on a subset of the training set. This subset can be chosen to be any size, but the common values are $5\%, 10\%$ or $25\%$ of the size of the entire training set. Compared to some more recent methods to reduce the number of evaluations needed to select an individual with lexicase selection \citep{ding2022scale, DeMelo2019LexSpeed}, down-sampling provides a sure-fire way to reduce the number of program executions needed as the size of the down-sample is chosen at will. These saved individual evaluations can be used to decrease runtimes, or can be used to evaluate more individuals and/or generations with the same fixed computational budget \citep{Hernandez:2019:subsampling, Helmuth2021benefits}. In this work, we will increase the number of generations such that the same number of individuals are evaluated as they would be when utilizing regular lexicase selection. For example, at a down-sampling rate of $0.1$, $(10\%$ of full training set), we increase the maximum generational limit by a factor of $10$ as this leads to the same number of individual evaluations per run as without performing down-sampling.

\citet{Hernandez:2019:subsampling}, \citet{Ofria:2019:GPTP} and more recently, \citet{Helmuth2021benefits} all find that down-sampled lexicase selection significantly improves the solution rate of GP runs across a variety of program synthesis benchmark problems. \citet{Helmuth2021benefits} find evidence for down-sampled lexicase selection's benefit being derived from a larger amount of program space being searched. Although it was a working hypothesis before that work, environmental change was ruled out as the likely cause of the benefits that down-sampled lexicase presents. To reach this conclusion, \citet{Helmuth2021benefits} compare down-sampled lexicase selection with \emph{truncated lexicase selection}, a variant of lexicase selection that allows for all cases to be used all generations, but simply cuts lexicase selection off after 10\% of the cases are used \citep{spector2017relaxations}. Note that the difference between truncated and down-sampled lexicase selection is that down-sampled samples the cases before randomly shuffling, and uses the same sample for every single selection in the generation. Truncated lexicase, on the other hand, shuffles and \emph{then} samples \emph{for every selection event}, meaning every individual selection event uses a possibly different set of cases. These experiments show that down-sampled lexicase does not improve on truncated lexicase selection, providing evidence for the fact that environmental change is not a driver of the success of down-sampled lexicase selection. We hypothesise that when using down-sampled lexicase selection, the lack of continuity of cases between generations might result in jarring discontinuities between the environments used in successive generations. These jarring discontinuities might result in members of the population struggling to adapt to difficult cases as they are taken out of the case pool before the population can gain a foothold on them. If true, this would be limiting the potential of down-sampled lexicase selection. In this work, we hope to study and potentially remedy this by causing \emph{incremental} environmental change, where there is some similarity between the down-samples so that individuals have more than one generation to adapt to the cases in current sample. In doing this, we explore whether or not down-sampled lexicase selection indeed does create jarring discontinuities between successive generations and what effects this has in practice.

\begin{table}[t]
    \centering
    \setlength{\tabcolsep}{16pt}
    \resizebox{0.8\columnwidth}{!}{
    \begin{tabular}{lr}
    \hline
    \textbf{Parameter}     & \textbf{Value} \\
    \hline
    runs per problem & 50 \\
    population size & 1000 \\
    training set size & 200 \\
    maximum generations & 300 \\
    variation operator & UMAD \\
    \end{tabular}
    }
    \caption{PushGP system parameters for the \emph{Fuel Cost} and \emph{Snow Day} problems. Note that in our experiments our maximum generation limit is set to different values depending on down-sampling rate.}
    \label{tab:pushGPparams}
\end{table}

\renewcommand{\arraystretch}{1.3} 
\begin{table*}[t]
\setlength{\tabcolsep}{12pt}
\centering
\begin{tabular}{lccccccc}
\hline
 Down-sampling Rate & \multicolumn{3}{c}{0.05} & \multicolumn{3}{c}{0.1} & 1\\
 \hline
 Method & \multicolumn{2}{c}{Rolling} & Random & \multicolumn{2}{c}{Rolling} & Random  & Lexicase \\
Downsample Size & 10 & 10 & 10 & 20 &  20  & 20 & 200\\
Step Size &  1  & 1 & 10  & 1 & 1 & 20 & 200      \\
Type &   Bag  & Queue    &  N/A &   Bag  &  Queue   & N/A & N/A \\
Max Gens &  6000   &   6000  &  6000  &   3000   &  3000  &  3000 & 300  \\
\hline
Successes &  41 & 40  & 39 & 35 & 36 & 40 & 17 \\
{\it p}-value vs Random & .80 & 1.00 & & .36 & .48
\end{tabular}
\caption{Successes out of 50 runs for the \emph{Fuel Cost} problem comparing rolling lexicase selection to simple random down-sampling and the baseline of full lexicase selection. Both rolling and random down-sampling significantly outperform the baseline. There are no significant differences between rolling and randomly down-sampled lexicase selection at the 0.05 level. These and subsequent {\it p}-values have been calculated using a pairwise chi-squared test to show the significance of the difference between the performance of the rolling methods compared to that of the control (random sampling) at the same down-sampling level.}
\label{tab:step1fc}
\end{table*}
\section{Methods}

The experiments in this paper were performed with the PushGP \citep{spector2005push3, spector2002genetic} framework. PushGP is a GP system that evolves computer programs written in the push programming language. The push programming language is a stack-based language that was designed specifically for genetic programming runs. It has the advantage of facilitating the evolution of programs that use multiple types and complex programming concepts such as conditional execution, recursion and iteration. For this paper, we use Propeller\footnote{https://github.com/lspector/propeller}, a Clojure implementation of PushGP. The PushGP system parameters that we used can be found in Table~\ref{tab:pushGPparams}.

The problems used in this paper come from the second program synthesis benchmark suite \citep{helmuth2021psb2}. This benchmark suite contains introductory programming problems that require programs to use a variety of data types and complex control flow structures. Specifically, the two problems we have chosen are \emph{Fuel Cost} and \emph{Snow Day} as these are problems where down-sampled lexicase selection has shown promise, but still has more room to improve. Programs that successfully solve \emph{Fuel Cost} must take a vector of positive integers, divide each by 3, round the result down to the nearest integer, and subtract 2. Then, the program must return the sum of all the new integers in the vector. The second problem, \emph{Snow Day}, requires solution programs to take an integer representing a number of hours and three floats representing how much snow is on the ground, the rate of snow fall, and the proportion of snow melting per hour. These solution programs must return the amount of snow on the ground after the amount of hours given.

We propose two different methods of down-sampling the training data for lexicase selection, where one is a direct response to experimental results from the other in hopes of understanding why the former was not successful. The first of these methods, \emph{rolling lexicase selection}, was an attempt to promote incremental environmental change to lexicase selection in hopes of scaffolding the evolution of PushGP programs by reducing the magnitude of discontinuities between down-samples in successive generations. The methods here describe experiments first comparing rolling lexicase selection with a step size of 1 to random down-sampling and full lexicase selection as a preliminary experiment. Then, an analysis on the effect of step size on solution rate was conducted on two problems in hopes of understanding the results of the first experiment. Finally, a new method, \emph{disjoint down-sampled lexicase selection} is proposed as an attempt to verify the effect that jarring discontinuities have on an evolving population of GP programs in practice.

\paragraph{Rolling Lexicase}
Rolling lexicase selection is a modification of down-sampled lexicase selection that iteratively changes the down-sample every generation, as opposed to all at once. This would have the effect of creating a multi-generational filter that does not have abrupt environmental changes like that for random down-sampling. We believe that the inclusion of incremental environmental change to this system where the environment is shifting would allow the members of the population to adapt more efficiently, driving larger success rates. To do this, we maintain a set of cases across generations in the active down-sample of cases. A new hyperparameter, the step size ($s$) is defined to be the number of cases we drop out of the current down-sample and replace with new cases from the entire training set. When this step size is equal to the down-sample size, this is exactly the same as performing down-sampled lexicase selection. Thus, this method can be thought of as a relaxation of down-sampled lexicase selection where we can choose the rate at which the environment changes. We also introduce two different versions of rolling lexicase, bag and queue, in hopes of understanding the differences a case being in the down-sample for a consistent amount of time would have. Using the bag method, the down-sample of cases are unordered and are dropped and added from the down-sample at random. \ie {}  if we need to drop out $s$ cases from the down-sample, these $s$ cases are chosen at random when using the bag method. This can be thought of as using a First In Random Out (FIRO) rolling strategy. On the other hand, the queue method of rolling lexicase selection offers a First In First Out (FIFO) rolling strategy. This means that the cases are removed from the down-sample in the order that they are added. Each case will therefore spend the exact same amount of time in the down-sample before being rolled out of it. On average, however, both of these methods of rolling would result in cases staying in the sample for roughly the same amount of time.

The rolling lexicase selection algorithm is outlined below. Note that this algorithm describes how the sample changes over evolutionary time as opposed to how a parent is selected. We use the lexicase selection algorithm as it is presented above to select our parents using the down-samples we have selected. We define $s$ to be the step size, $d$ the down-sampling rate, $N$ the training set size, $n=N\times d$ the down-sample size, and $g = \frac{G}{d}$ the generational limit (where $G$ is the set generational limit for regular lexicase selection).
\begin{enumerate}
    \item \textbf{candidates} is set to initially contain the entire population.
    \item \textbf{cases} is set to contain the entire training set.
    \item \textbf{case-sample} is set to initially contain a random sample of size $n$ from \textbf{cases}.
    \item Until solution found or the generational limit $g$ is reached:
    \begin{enumerate}
        \item Evaluate all individuals on the cases in \textbf{case-sample}, generating individual error vectors of length $n$.
        \item If any individuals pass all the cases in \textbf{case-sample}, re-evaluate the best individual on the cases in \textbf{cases}. If these are all passed as well, a candidate solution has been found. Re-evaluate the best individual on the test set. If this individual passes all of these cases, this counts as a success.
        \item Else, using these error vectors, use lexicase selection to select a set of parents.
        \item Apply variational operators on these parents to produce the next population.
        \item if using the bag variety of rolling, remove $s$ random cases from \textbf{case-sample}, and add $s$ new ones from \textbf{cases} (that are not already used). If using the queue variety of rolling, dequeue $s$ random cases from \textbf{case-sample}, and enqueue $s$ new ones from \textbf{cases}.
    \end{enumerate}
\end{enumerate}

\paragraph{Disjoint Lexicase}
Disjoint down-sampled lexicase selection is proposed as a method that does close to the opposite of rolling lexicase selection. As opposed to maintaining a small set of cases across generations, disjoint down-sampled lexicase selection ensures that not only are cases not maintained across generations, they are not even allowed to re-enter the down-sample until every other case has been used up once. This can be thought of (and indeed how it is implemented) as first selecting $n$ random partitions of the training set of size $N$, where each partition is of size $\frac{N}{n}$, and picking one of these partitions every generation until they run out. When this happens, re-partition the training data, and repeat. With this treatment, cases will not be repeated for a number of generations. We predict that this method will take the jarring discontinuities between down-samples to a more extreme level than that with simple random down-sampling. By doing this, we hope to see whether label-wise discontinuities (the differences between the value of the input and output cases) in practice result in real discontinuities in which behaviors are selected for across generations. We compare this variant of down-sampling to rolling and randomly down-sampled lexicase selection to explore the effects a more jarring environmental change might have on the evolution of solutions to GP problems.

\section{Results and Discussion}

\paragraph{Preliminary Experiment}
First, we ran a comparison of rolling lexicase with a step size of 1, random down-sampled lexicase, and full lexicase selection on the \emph{Fuel Cost} problem. This experiment was meant to be a preliminary exploration into incrementally varying the environment for GP runs. The results from these runs can be found in Table~\ref{tab:step1fc}.

The results from this experiment lead us to believe that there was no significant improvement when incrementally shifting the environment for down-sampled lexicase selection. Although not statistically significant, the relative success rates between rolling and down-sampled lexicase at the down-sampling levels of 0.05 and 0.1 suggested that the proportion of cases in the down-sample that are rolled might be important to consider. At the 0.05 down-sampling rate, cases are entirely refreshed (approximately for the bag method) every 10 generations. When using a 0.1 down-sampling rate, this number doubles to 20. The trend here seems to suggest that the longer it takes to refresh the training set, the lower success rates become. In order to explore the effect the rate of environmental change has on the evolution of GP solutions, we conduct experiments varying the step sizes for rolling lexicase selection across two different program synthesis problems. This experiment is outlined in the next section.

It is also interesting that we did not observe any differences between the bag and queue methods of maintaining a sample for rolling lexicase selection. This might be due to the fact that the individuals might require multiple generations to adapt to certain hard cases, while other cases are passed by a lot of individuals and are no longer informative. Since we are randomly selecting the next case to be added in, which could be at any level of importance for the population, any regularity in rolling with the queue method would not result in any meaningful differences to simply randomly rolling. Dropping out the case that was added first might simply have the same effect as dropping out a random case, as the only time this dropping procedure significantly affects the population is if it drops out an important case, which are placed randomly throughout the down-sample. In short, the difference between using a bag and a queue seems to be negligible when rolling due to different cases measuring similar things and the fact that cases will be around for the same amount of time on average.


\begin{table}[t]
\centering
\resizebox{\columnwidth}{!}{
\begin{tabular}{lcccccc}
\hline
 Method & \multicolumn{5}{c}{Rolling} & Random\\
 \hline
 Step Size{ }  & 1 & 3 & 5 & 10 & 19 & 20\\
 \hline
 Successes & 35 & 32 & 32 & 36 & 36 & 40\\
 {\it p}-value & .35 & .12 & .12 & .48 & .48 & 
\end{tabular}
}
\caption{Rolling at different rates. Effect of step size on success rate for the \emph{Fuel Cost} problem, using the bag variety of rolling down-sampled lexicase selection. {\it p}-values are shown to quantify the significance of the difference between each rolling method and purely random down-sampling.}
\label{tab:fuelVaryStep}
\end{table}

\begin{table}
\centering
\setlength{\tabcolsep}{8pt}
\resizebox{\columnwidth}{!}{
\begin{tabular}{lcccccc }
\hline
 Method & \multicolumn{4}{c}{Rolling} & Random\\
 \hline
 Step Size{ } & 2 & 5 & 10 & 19 & 20\\
 \hline
 Successes & 23 & 20 & 22 & 18 & 26\\
 {\it p}-value & .60 & .30 & .54 & .16 & & 
\end{tabular}
}
\caption{Rolling at different rates. Effect of step size on the success rate for the \emph{Snow Day} problem, using the bag variety of rolling down-sampled lexicase selection. Although not important for this experiment, regular lexicase selection achieves $7$ successes out of $50$ runs.}
\label{tab:snowVaryStep}
\end{table}

\begin{table*}[t]
\centering
\setlength{\tabcolsep}{12pt}
\begin{tabular}{lcccccc@{\hskip 0.35in}c}
\hline
 Down-sampling Rate & \multicolumn{2}{c}{0.01} & \multicolumn{2}{c}{0.05} & \multicolumn{2}{c}{0.1} & 1\\
  Down-sampling Type & Random & Disjoint & Random & Disjoint & Random & Disjoint & Lexicase\\
 Down-sample Size & 2 & 2 & 10 & 10 & 20 & 20 & 200 \\
 Max-Gens & 30000 & 30000 & 6000 & 6000 & 3000 & 3000 & 300 \\
 \hline
 Successes & 33 & 33 & 39 & 42 & 40 & 33 & 17\\
 {\it p}-value vs Random & & 1.00 & & .61 && .18

\end{tabular}
\caption{A comparison of disjoint down-sampled lexicase selection, simple random down-sampled lexicase selection, and regular (full) lexicase selection on the \emph{Fuel Cost} problem across a range of down-sampling rates. Using disjoint samples as opposed to random down-sampling results in no statistically significant difference in success rates on this problem. All runs using random or disjoint lexicase selection result in a significant improvement on full lexicase selection.}
\label{tab:fuelDisjoint}
\end{table*}

\begin{table}[t]
\centering
\setlength{\tabcolsep}{6pt}
\begin{tabular}{lcc@{\hskip 0.35in}c}
\hline
 Down-sampling Type & Random & Disjoint &  Lexicase\\
 Down-sample Size & 20 & 20 & 200 \\
 Max-Gens & 3000 & 3000 & 300 \\
 \hline
 Successes & 26 & 19 & 7\\

\end{tabular}
\caption{A comparison of disjoint lexicase selection to random down-sampling at the 0.1 down-sampling rate on the \emph{Snow Day} problem. As a control, the regular lexicase selection success rate is shown on the right. There are no statistically significant differences between the success rates when using random down-sampling and disjoint down-sampling ({\it p}=.23). Both random and rolling down-sampled lexicase selection significantly improve on (full) lexicase selection (${\it p}<0.02$).}
\label{tab:snowDisjoint}
\end{table}

\paragraph{Step Size Variations}
In order to test our reasoning for the preliminary experiment's negative results, we attempt to vary the evolutionary time it takes for a down-sample to be entirely refreshed. To do this, we repeated the preliminary experiments with step sizes other than 1 for the \emph{Fuel Cost} and \emph{Snow Day} problems. To keep the comparisons consistent, we used a down-sampling rate of 0.1, and the bag method for rolling (as the queue method did not perform differently to the bag method). The \emph{Fuel Cost} results can be found in Table~\ref{tab:fuelVaryStep}, and the \emph{Snow Day} results can be found in Table~\ref{tab:snowVaryStep}. These results further reinforce the claim that rolling lexicase selection is not significantly better than simple random downsampling. No individual run has a statistically significant difference in performance to random down-sampling, but there might be a low magnitude signal when considering multiple runs using rolling lexicase at different steps sizes across two different problems. All 9 of the rolling lexicase runs performed worse than the randomly down-sampled run did.
 In fact, most of the runs have a handful fewer successes. This leads us to believe that, despite our intuition and biological inspiration, random incremental environmental change seems to not only \emph{not} be an improvement on simple random down-sampling, but it might even be worse. While we believed that allowing for incremental environmental change would create a multi-generational filter that does not have jarring discontinuities and would provide for evolutionary advantages, it seems like achieving this ends through randomly rolling the cases is not an effective strategy.  Whilst not that strong of a trend, it seems possible that the runs that have a lower step size perform worse than those at a higher step size.
 

A possible reason for the neutral or negative effect rolling lexicase selection has is that cases that represent certain niches might be left out for a long time (as it takes so long to go over the entire training set with a small step size). Say, for example, that we were evolving a program to perform division. The test case representing a division by zero is obviously a very important case to consider for the selection of a candidate solution. When using full lexicase selection, this case always has a chance to be placed near the beginning of a shuffle, and therefore will always have a chance to exert selection pressure. When down-sampling, this case might be left out for a few generations, but will continually cycle in every once in a while. When rolling, however, this case might be out of the current case pool for many generations, which might be long enough for the individuals to catastrophically forget how to divide by zero. This would result in certain niches getting closed out of the population, resulting in slight evolutionary performance losses. There is an intricate trade-off between having a case around for long enough for the population to adapt to it and not having the other cases out of the down-sample for too long that the population forgets how to solve them. For the \emph{Fuel Cost} problem, the step sizes of 3 and 5 seem to perform worse than the rest, which is possibly the place where this trade-off is not made optimally. 

We also believe that the prevalence of synonymous cases plays an extremely important role in the above comparisons. Synonymous cases are cases in the training set that do not contain exactly the same labels, but measure very similar behavior. These can be thought of as a set of cases that would be passed by a very similar set of individuals in the population every generation. This means that two sets of cases that are entirely different from each other when it comes to input and output labels could result in the selection of the very same set of parents. It could be that the existence of these cases would further dilute the effect rolling would have on down-sampled lexicase selection as a randomly picked set of cases might result in the exact same individual being selected to that if we maintained a few cases from the last generation. In practice, we expect that the cases used in our program synthesis problems are somewhat or even highly synonymous, with many cases testing the same behavior of programs in very similar ways. This would explain the neutral or negative effect that rolling seems to have on our down-sampled lexicase selection runs at a variety of step sizes.

\paragraph{Disjoint Samples}
Due to the lack of success of rolling the down-samples between generations, we test whether going in the opposite direction through \emph{disjoint} samples would have a significant effect on the success rate of our GP runs. If randomly down-sampling the training data does indeed result in jarring discontinuities between generations, this method of down-sampling would only exacerbate the effect of doing so.  To test this, we performed experiments comparing lexicase selection with random down-sampling and disjoint down-sampling on the \emph{Fuel Cost} and \emph{Snow Day} Problems. Those results can be found in Table~\ref{tab:fuelDisjoint} and \ref{tab:snowDisjoint}.


This set of results shows that using disjoint down-samples is likely not significantly different to regular random down-sampling. While going against our prior intuition, it seems that these disjoint samples would not significantly change the way selection works. A likely reason for these results could also be due to the prevalence of synonymous cases in our training set, which would mean that two samples that are entirely disjoint might end up being very similar when it comes to which parent is selected using them. Although disjoint down-sampling does not allow \emph{the exact same} case to be in temporally close generations, these synonymous cases can and ultimately do end up in successive generations. This would result in no significant difference between disjoint down-sampling and random down-sampling, as supported by our empirical results.

It is also possible that disjoint down-sampling is performing worse than random when we increase the size of the down-sample to 20. A possible reason for this could be that when the down-samples are big enough, all cases that are synonymous could be randomly placed in the same down-sample, resulting in similar catastrophic forgetting to that which we hypothesize could have happened when rolling. For example, if all 5 cases that represent the same case are placed in the same down-sample, this entire case niche will not be used for the next 9 generations, which could be enough time for the population to forget how to solve this case.

\section{Conclusion and Future work}
In this paper, we investigate the hypothesis that down-sampled lexicase selection causes discontinuities between the generations of program synthesis GP runs. To do this, we present two new methods of down-sampling the training data for lexicase selection: \emph{rolling (down-sampled) lexicase selection} and \emph{disjoint (down-sampled) lexicase selection}. Rolling lexicase selection is an attempt to encourage incremental environmental change in order to better allow the population to adapt to their changing environment by reducing the amount of jarring discontinuities between consecutive generations. Through an experiment where we only vary the down-samples by one case every generation, we find that incrementally changing environments randomly seems to not improve on regular random down-sampled lexicase selection. To test whether we were rolling too slowly, we repeated the experiment, varying the step size. We find that no intermediate step size significantly outperforms down-sampled lexicase selection either.

In order to investigate the reasons that rolling lexicase selection was not as efficacious as we expected, we performed a third set of experiments using disjoint samples. This method of down-sampling the training set was designed to take the discontinuities that we predicted exist when randomly down-sampling to the extreme. To do this, we propose \emph{disjoint lexicase selection}, whereby the set of training cases are split into disjoint sets, and each set is used for one generation. Once all cases are used once, we re-partition the cases, allowing for the cases to be used for a second time. Through these experiments, we find that disjoint down-sampling does not significantly affect the solution rates when compared to random down-sampling for lexicase selection. While using purely disjoint samples would result in forced jarring discontinuities when considering the input and output labels of the cases, we hypothesize that this does not indeed result in jarring discontinuities in the \emph{behaviors} being measured due to the large presence of synonymous cases. When the down-samples are large enough, it is possible that each down-sample could include all synonymous cases that measure the same behavior. This could lead to slightly reduced performance when using disjoint down-sampled lexicase selection due to the population forgetting how to perform that behavior (because that case-niche is not used for multiple generations). 

Despite the fact that these results do not present a way to improve success rates through environmental change in lexicase selection, we believe that this work highlights a promising direction for future exploration. As opposed to randomly down-sampling the training cases, perhaps a more intelligent approach would result in evolutionary runs that benefit from lower computational costs, without the loss of information from the omission of some vital cases in the training set. In particular, we believe sampling methods that take into account case synonymy, unlike rolling and disjoint samples, may lead to better results. This could perhaps be achieved by the dynamic collection of population statistics regarding the interaction between the individuals and the training cases. These statistics would then be used to select down-samples that are likely to maintain behavioral niches of individuals, while moderating the effort put into synonymous cases. 


\section{Acknowledgements}
This material is based upon work supported by the National Science Foundation under Grant No. 1617087. Any opinions, findings, and conclusions or recommendations expressed in this publication are those of the authors and do not necessarily reflect the views of the National Science Foundation.

This work was performed in part using high performance computing equipment obtained under a grant from the Collaborative R\&D Fund managed by the Massachusetts Technology Collaborative.

The authors would like to thank Charles Ofria, Alexander Lalejini, Jose Guadalupe Hernandez, Edward Pantridge, Anil Saini and Li Ding for discussions that helped shape this work.

\footnotesize
\bibliographystyle{apalike}
\bibliography{example} 

\end{document}